\documentclass[runningheads]{llncs}
\usepackage{graphicx}
\usepackage{xcolor}
\usepackage{comment}
\usepackage[utf8]{inputenc}
\usepackage{multirow}
\usepackage{arydshln}
\usepackage{subfig}
\usepackage{float}
\usepackage{array}
\usepackage{tabularx}
\usepackage{graphicx}
\usepackage{url}
\usepackage{adjustbox}
\usepackage{xspace}
\usepackage{xcolor}
\usepackage{comment}
\usepackage{caption} 
\usepackage{booktabs}
\usepackage{hyperref}
\PassOptionsToPackage{bookmarks=false}{hyperref}
\usepackage{nicematrix}
\restylefloat{table}
\usepackage{latexsym}
\usepackage{amsmath, amssymb}
\usepackage{type1cm}        
\usepackage{makeidx}         
\usepackage{graphicx}        
\usepackage{multicol}        
\usepackage[bottom]{footmisc}

\usepackage{newtxtext}       %
\usepackage[varvw]{newtxmath} 
\usepackage{type1cm}        
\usepackage{makeidx}         
\usepackage{graphicx}        
\usepackage{multicol}        
\usepackage[bottom]{footmisc}

\usepackage{newtxtext}       %
\usepackage[varvw]{newtxmath}       

\begin{document}
\mainmatter 
\title{How Green are Neural Language Models?\\ Analyzing Energy Consumption in Text Summarization Fine-tuning}
\titlerunning{How Green are Neural Language Models?}  
%
\author{Tohida Rehman\inst{1} \and Debarshi Kumar Sanyal\inst{2} \and Samiran Chattopadhyay\inst{3}}
\authorrunning{Rehman, Sanyal, and Chattopadhyay} 
%
\tocauthor{Tohida Rehman, Debarshi Kumar Sanyal, Samiran Chattopadhyay}
\institute{Jadavpur University, Salt Lake Campus, Kolkata 700106, India,\\
\email{tohidarehman.it@jadavpuruniversity.in},\\
\and
Indian Association for the Cultivation of Science, Jadavpur 700032, India,\\
\email{debarshi.sanyal@iacs.res.in},\\
\and
Techno India University, Salt Lake, Kolkata 700091, India,\\
\email{samirancju@gmail.com}
}

\maketitle              
\begin{abstract}
Artificial intelligence systems significantly impact the environment, particularly in natural language processing (NLP) tasks. These tasks often require extensive computational resources to train deep neural networks, including large-scale language models containing billions of parameters. This study analyzes the trade-offs between energy consumption and performance across three neural language models: two pre-trained models (T5-base and BART-base), and one large language model (LLaMA-3-8B). These models were fine-tuned for the text summarization task, focusing on generating research paper highlights that encapsulate the core themes of each paper. The carbon footprint associated with fine-tuning each model was measured, offering a comprehensive assessment of their environmental impact. It is observed that LLaMA-3-8B produces the largest carbon footprint among the three models. A wide range of evaluation metrics, including ROUGE, METEOR, MoverScore, BERTScore, and SciBERTScore, were employed to assess the performance of the models on the given task. This research underscores the importance of incorporating environmental considerations into the design and implementation of neural language models and calls for the advancement of energy-efficient AI methodologies.
\end{abstract}

\section{Introduction}
\label{intro}
The environmental impact of artificial intelligence (AI) systems, particularly in natural language processing (NLP), has become a pressing concern in recent years. NLP tasks, such as text summarization, question-answering, machine translation, and named-entity recognition often require the use of deep neural networks with a large number of parameters. The recent trend is to use pre-trained language models (PLMs) and large language models (LLMs) for these tasks as they produce excellent performance \cite{minaee2402large}. Though PLMs and LLMs are both pre-trained in a self-supervised manner, a PLM typically has fewer than a billion parameters while an LLM is larger. These models require extensive training on billions to trillions of tokens, spanning multiple GPUs and several days. This training process consumes huge energy, leading to significant environmental implications. These pre-trained models are then fine-tuned for specific tasks, which again produces high consumption. Despite the advancements in AI, the computational costs of fine-tuning these models for specific tasks have yet to be fully explored, especially in terms of their energy efficiency and environmental sustainability.

The application we focus on in this paper is summarization; in particular, generation of \textit{research highlights} from scientific papers. Because of the rapid increase in scientific publications over the last few decades \cite{bornmann2021growth}, it has become challenging for researchers to keep up with the latest developments. To address this, many publishers have introduced research highlights, which are brief summaries of key findings, that accompany the abstract. These highlights are easier to understand, particularly on mobile devices, and improve search engine indexing and content discovery. However, not all research papers include these highlights, and it would be useful to be able to generate them automatically. Text summarization, which helps reduce the time needed to extract crucial information from large documents, can be done through extractive or abstractive methods. Extractive summarization selects relevant sentences directly from the text, while abstractive summarization generates novel words and sentences to convey the core message, typically yielding more coherent and insightful summaries \cite{luhn1958automatic}. Research highlights are typically abstractive summaries. 

In this paper, we study the problem of generation of research highlights given the abstract of a paper. However, our investigation will not only include fine-tuning and evaluating pre-trained language models but also an empirical analysis of the energy consumption of fine-tuning. 
As the size and complexity of these models increase, so do their computational costs, leading to greater environmental concerns. We analyze the energy requirements of two PLMs and an LLM (all  with open weights), with the aim of understanding the balance between their performance and energy efficiency. We have made a demo of our highlight-generation system publicly  available\footnote{\url{https://highlightsgeneration-mixsub.onrender.com}}. By quantifying the energy usage involved in fine-tuning these models, we aim to emphasize the need for more sustainable practices in the development of NLP technologies.

The main contributions of this work are as follows:
\begin{enumerate}
    \item We fine-tuned three distinct language models: \textbf{T5-base}, \textbf{BART-base}, and \textbf{LLaMA-3-8B}, adapting them specifically for the task of generating research highlights that encapsulate the core findings of a paper.
    \item We assessed the performance of the models through various metrics, including \textbf{ROUGE}, \textbf{METEOR}, \textbf{MoverScore}, \textbf{BERTScore}, and \textbf{SciBERTScore}, to evaluate their effectiveness in the given task.
    \item We carried out an extensive analysis of the energy consumption and carbon emissions during the fine-tuning process of each model, providing a comprehensive view of their environmental impact in NLP research.
    \item Our findings reveal the trade-offs between model size, computational resources, performance, and energy consumption, offering meaningful insights into the sustainability of cutting-edge NLP technologies.
\end{enumerate}

\section{Literature Survey}
\label{literature}
Transformer-based neural language models \cite{vaswani2017attention} have substantially advanced the performance of NLP tasks, such as text summarization and question-answering. However, the increasing complexity and scale of these models have raised significant concerns regarding their environmental impact, particularly in terms of energy consumption.
Strubell et al. \cite{strubell2019energy} conducted one of the earliest evaluations of the carbon footprint associated with training large-scale pre-trained language models, including transformer-based architectures \cite{vaswani2017attention} such as BERT. Their investigation demonstrated that the training of cutting-edge NLP models, particularly when utilizing extensive datasets and prolonged training phases, can result in significant CO$_2$ emissions. This study emphasized the urgency of developing approaches to minimize both the energy demands and the environmental impact of training NLP models. 
Patterson et al. \cite{patterson2021carbon} examined the trade-off between model size and energy efficiency, showing that while larger models improve performance, they also significantly increase energy consumption, especially for resource-demanding tasks like text summarization.
Schwartz et al. \cite{schwartz2020green} called for the adoption of `Green AI' practices, urging the development of frameworks to evaluate and minimize the environmental impact of AI systems. They stressed the need to integrate environmental factors into the design and assessment of machine learning models, ensuring that progress in AI does not lead to excessive ecological costs. Recently, researchers have conducted comprehensive surveys of `Green AI' solutions proposed in the literature \cite{verdecchia2023systematic,alzoubi2024green}.

Automatic text summarization has evolved over time, beginning with Luhn et al. \cite{luhn1958automatic}, who introduced an extractive approach for summarizing technical papers by selecting sentences based on word frequency, excluding common terms. The field advanced further with sequence-to-sequence models, attention-based encoders, and pointer-generator networks \cite{See2017GetTT}, which not only handled out-of-vocabulary (OOV) words but also reduced repetitive phrase generation, as shown in \cite{rehman2021automatic,rehman-etal-2022-named,rehman2023research,10172215}. The introduction of the transformer architecture \cite{vaswani2017attention} marked a significant shift, leading to the development of pre-trained models like T5 \cite{JMLR:v21:20-074}, BART \cite{lewis-etal-2020-bart} and many more. These models are trained on large, general-purpose datasets and can be fine-tuned for specific tasks. Rehman et al. \cite{rehman2022analysis} performed a comprehensive evaluation of pre-trained models, including google/pegasus-cnn-dailymail, T5-base, and facebook/bart-large-cnn. Their study assessed the models' performance in text summarization tasks using various datasets, such as CNN-DailyMail, SAMSum, and BillSum.

The task of generating research highlights, however, has been studied only recently. Collins et al. \cite{collins2017supervised} employed supervised learning and a binary classifier to identify relevant highlights from text. Cagliero and Quatra \cite{cagliero2020extracting} employed multivariate regression to identify the top-$k$ most pertinent sentences. Rehman et al. \cite{rehman2021automatic} proposed an abstractive method utilizing pointer-generator networks to generate research highlights from abstracts, and later enhanced this approach by incorporating named entity recognition \cite{rehman-etal-2022-named}. Rehman et al. \cite{10172215} also conducted a comprehensive study on highlight generation using various deep learning models.

\section{Datasets}
\label{datasets}
We have used the \textbf{MixSub} dataset \cite{10172215} for the highlight generation task. The dataset was created by collecting research articles from ScienceDirect\footnote{\url{https://www.sciencedirect.com/}}. It includes 19,785 research articles from multiple domains, published in journals in 2020, with each article paired with its corresponding author-written research highlights. Each entry consists of an \textit{abstract} and   \textit{highlights}. The dataset is  partitioned into training, validation, and test sets in an 80:10:10 ratio. For the experiments, we utilized 5,000 samples from the training set, 625 samples from the validation set, and 625 samples from the test set. An example from the MixSub dataset is shown in Figure \ref{fig:sample_MixSub}.
\begin{figure*}[!htbp] 
\centering 
\begin{tabular}{ |p{12cm}|} \hline
{\bf Abstract:} \colorbox{blue!10}{``We use a controlled experiment to analyze the impact of watching} \colorbox{blue!10}{different types of educational traffic campaign videos on overconfidence} of undergraduate university students in Brazil. The videos have the same underlying traffic educational content but differ in the form of exhibition. \colorbox{pink!30}{We find that} \colorbox{pink!30}{videos with shocking content (Australian school) are more effective in reducing} \colorbox{pink!30}{drivers’ overconfidence}, followed by those with punitive content (American school). \colorbox{yellow!20}{We do not find empirical evidence that videos with technical}  \colorbox{yellow!20}{content (European school) change overconfidence.} Since several works point to a strong association between overconfidence and road safety, our study can support the conduit of driving safety measures by identifying efficient ways of reducing drivers' overconfidence. Finally, \colorbox{green!10}{this paper also introduces how to use machine learning techniques to mitigate the} \colorbox{green!10}{usual subjectivity in the design of the econometric specification} that is commonly faced in many researches in experimental economics.''
\\\hline 	    
{\bf Author-written research highlights:} 
\begin{itemize}
    \item[$\blacktriangleright$] \colorbox{blue!10}{``We use a controlled experiment to analyze the impact of watching different} \colorbox{blue!10}{types of educational traffic campaign videos on overconfidence.''} 
    \item[$\blacktriangleright$] \colorbox{pink!30}{``We find that videos with shocking content (Australian school) are more} \colorbox{pink!30}{effective in reducing drivers overconfidence.''} 
    \item[$\blacktriangleright$] \colorbox{yellow!20}{``We do not find empirical evidence that videos with technical content} \colorbox{yellow!20}{(European school) change overconfidence.''} 
    \item[$\blacktriangleright$] \colorbox{green!10}{``This paper also introduces how to use machine learning techniques to} \colorbox{green!10}{mitigate the usual subjectivity in the design of the econometric specification.''}
\end{itemize}\\\hline
\end{tabular} 	
\caption{\small An (abstract, highlights) pair from the MixSub dataset.  \texttt{\url{https://www.sciencedirect.com/science/article/pii/S0001457519307213}}. We have used colors to denote the correspondence between a highlight and the abstract. Here, we find each highlight to be a segment of some sentence in the abstract. However, \textit{not} in all papers, is there a straightforward mapping from abstracts to sentences in the abstract. In particular, a highlight may combine information from multiple sentences in the abstract and express it using  very different words.}	
\label{fig:sample_MixSub} 
\end{figure*}

\section{Models Used}
In this section, we describe the different pre-trained Language Models (PLMs) used in our study. The models are as follows:

\begin{enumerate}
    \item \textbf{T5-base} \cite{JMLR:v21:20-074}: Built on an encoder-decoder framework, this model is a refined adaptation of the transformer architecture introduced by Vaswani et al. \cite{vaswani2017attention}. It unifies diverse text processing tasks, such as translation, question answering, and classification, under a single framework by representing them as `text-to-text' transformation problems. During pre-training, the model is trained to reconstruct corrupted or masked spans of text. T5-base contains 220 million parameters.
    
    \item \textbf{BART-base} \cite{lewis-etal-2020-bart}: BART is a denoising autoencoder that combines bidirectional and autoregressive transformers, drawing inspiration from BERT and GPT, respectively. During pre-training, noise is introduced into the input text using a corruption function, and the model learns to reconstruct the original content. This architecture proves highly effective for text generation tasks. BART-base contains 139 million parameters.
     
    \item \textbf{LLaMA-3-8B} \cite{touvron2023llama}: We have used the pre-trained LLaMA-3-8B model\footnote{\url{https://ai.meta.com/blog/meta-llama-3/}}, which contains 8 billion parameters. The LLaMA models \cite{touvron2023llama} are designed using a decoder-only Transformer architecture and were trained exclusively on publicly available datasets, setting them apart from the GPT models.
\end{enumerate}

\section{Performance Evaluation Method}
To evaluate the quality of the generated outputs for research highlight generation, we utilized standard automatic text summarization evaluation metrics. These include ROUGE , METEOR, MoverScore  BERTScore, and SciBERTScore \cite{rehman2025can}. 
ROUGE-$n$ measures the overlap of $n$-grams between the generated output and the reference highlights. We have used ROUGE-1 (unigrams), ROUGE-2 (bigrams), and ROUGE-L (longest common subsequence). METEOR evaluates sentence-level alignment between the generated and reference highlights. 

However, ROUGE and METEOR are limited in their ability to judge the semantic similarity between model-generated and ground-truth results, especially when the model generates novel words to express the same information. Hence, for improved semantic similarity evaluation, we have used MoverScore, BERTScore, and SciBERTScore. MoverScore combines Word Mover's Distance and contextual embeddings to evaluate the semantic alignment of the generated and reference outputs. BERTScore computes the cosine similarity of BERT embeddings between the generated and reference texts. Since our dataset contains scientific papers, we experimented with SciBERT \cite{beltagy2019scibert} in place of BERT to calculate the BERTScore, and call the modified metric SciBERTScore. Recollect that SciBERT has been pre-trained on documents in the computer science and biomedical fields, and therefore captures more domain-specific information compared to BERT which was pre-trained on general-domain texts. 

\section{Evaluation of Energy Usage}
The proliferation of transformer-based language models and other parameter-heavy deep neural models has led to large computational pre-training and fine-tuning phases. This produces a formidable carbon footprint. Researchers have developed models to measure the carbon emission   \cite{strubell2019energy,lannelongue2021green,bouza2023estimate}.
We follow a widely used approach, proposed in  \cite{lannelongue2021green}, to estimate the carbon footprint $C$ of training a model, in \textit{grams of carbon dioxide equivalent}, or gCO$_2$e, as outlined in Equation \ref{eq:eq_carbon_footprint_lanne}:
\begin{equation}
\label{eq:eq_carbon_footprint_lanne}
C = t\times(n_c\times P_c \times u_c + n_m \times P_m) \times PUE \times CI \times 0.001 
\end{equation}
In the above, $t$ represents the running time (in hours), $n_c$ denotes the number of cores (CPU or GPU), $P_c$ indicates the power consumption of a computing core, $u_c$ refers to the core usage factor (ranging from 0 to 1), $n_m$ is the amount of memory available (in gigabytes) and $P_m$ is the power consumption of a memory unit (in watts). $P_m$ is assumed to be 0.3725 W/GB, as reported in \cite{lannelongue2021green}. Power Usage Effectiveness or $PUE$ is the efficiency coefficient of the data center. PUE is defined as the ratio between the total power drawn by the data center facility and the power delivered to the computing equipment. While $PUE = 1$ is the ideal case, in practice, $PUE > 1$ as some energy is consumed by non-computing devices in the facility. CI denotes the Carbon Intensity (measured in gCO$_2$e/kWh), which is defined as the carbon footprint of producing 1 kWh of energy. CI varies between locations due to variations in production methods. The global average carbon intensity (CI) is 475 gCO$_2$e/kWh \cite{CarbonIntensity}.

To identify the CPUs and GPUs explicitly in the set of computing cores, we rewrite Equation \ref{eq:eq_carbon_footprint_lanne} as follows:
\begin{align}
\label{eq:Our_carbon_footprint}
C &=  t \times (n_{cpu}\times P_{cpu} \times u_{cpu} + n_{gpu} 
   \times P_{gpu} \times u_{gpu}  + n_m \times P_m) \nonumber \\
   & \times PUE \times CI \times 0.001
\end{align}
where $n_{cpu}$ denotes the number of CPU cores, $P_{cpu}$ indicates the power consumption of a CPU core, $u_{cpu}$ refers to the CPU core usage factor (ranging from 0 to 1), $n_{gpu}$ is the number of GPU cores, $P_{gpu}$ is the power consumption of the GPU core, $u_{gpu}$ is the GPU usage factor (also between 0 and 1).

\section{Experimental Setup}
In this section, we discuss the data pre-processing procedures and the implementation details.
During the preprocessing stage, we began by eliminating excess whitespace from the documents and selecting only those samples where the abstract consisted of a minimum of 20 tokens, and the paper highlights contained at least 3 tokens. We set a limit of 100 tokens for the generated highlights and 512 tokens for the input abstracts.
We selected the two PLMs from the HuggingFace model hub: T5-base \footnote{\url{https://huggingface.co/t5-base}} and BART-base \footnote{\url{https://huggingface.co/facebook/bart-base}}. These models were fine-tuned for 5 epochs with a batch size of 16 and a learning rate of 2e-5. For the task of generating research highlights using the LLaMA-3-8B model \footnote{\url{meta-llama/Meta-Llama-3-8B}} from the HuggingFace, we employed the following prompt during training:\\
\texttt{Create a concise highlight from this abstract using no more than 100 tokens, focusing on the main contributions and key points. <ABSTRACT>}\\
LLaMA-3-8B was fine-tuned for 3 epochs with the parameter-efficient technique called Low-Rank Adaptation (LoRA) \cite{hu2022lora} with a learning rate of 1e-4. Both training and evaluation were carried out using a batch size of 16. The hyperparameters for LoRA were set as follows: {\tt rank = 128}, {\tt lora\_alpha = rank * 2}, and {\tt lora\_dropout = 0.05}.

\section{Results}
\label{sec:results}

\subsection{Performance Comparison of the Fine-Tuned Models}
\label{sec:quanResults}
We evaluated the models on the MixSub dataset using ROUGE, METEOR, MoverScore, BERTScore, and SciBERTScore. The results in Table \ref{Table:par_all_types_rouge_meteor_bert-mixsub} show that T5-base and BART-base perform similarly in ROUGE and METEOR, outperforming LLaMA-3-8B. However, when using semantic metrics like MoverScore, BERTScore, and SciBERTScore, the models’ scores are closer. This indicates that LLaMA-3-8B generates outputs semantically similar to that of the other models, but with different wording and therefore, lower lexical overlap scores. 

\begin{table*}[!h]
\centering
\caption{\small Evaluation of generated highlights for \textbf{MixSub}.}
\label{Table:par_all_types_rouge_meteor_bert-mixsub}
\begin{adjustbox}{width=1.0\linewidth}
{\begin{tabular}{|lccccccc|} \hline
Model Name &ROUGE-1 &ROUGE-2 &ROUGE-L &METEOR  &MoverScore &BERTScore &SciBERTScore\\\hline
T5-base &32.91 &\textbf{12.08} & \textbf{23.6} & \textbf{29.94} &17.76  &86.32 &63.7 \\ \hline
BART-base & \textbf{34.28} &11.98 &22.87 &28.81 & \textbf{20.05}  & \textbf{87.0} & \textbf{65.42} \\ \hline
LLaMA-3-8B &27.41 &7.71 &19.2 &23.73 &18.44  &86.84 &65.02\\ \hline

\end{tabular} }
\end{adjustbox}
\end{table*}

Figure \ref{fig:sample_Abs_Highlights_MixSub} presents a case study. Although the highlights generated by T5-base and BART-base are closer to the author-written highlights than LLaMA-3-8B, they contain repetitive information while LLaMA-3-8B's output is  cleaner. Probably more examples are needed to align the latter's output to the author-written text. Alternatively, a fine-tuned LLM like LLaMA or ChatGPT could be used in a zero-shot setting to perform linguistic correction of the output of the smaller models.
\begin{figure*}[!htbp] 
 \centering 
 \begin{tabular}{|p{12.5 cm}|} \hline
 {\bf Author-written highlights:}``Inspired by the self organizing aggregation behavior observed in nature we developed a fuzzy logic based self organizing aggregation method for application in swarm robots. The fuzzy logic approach has increased the coordination between the swarm robots and provided the transition from individual robot movement to collective swarm movement. The performance and scalability of the developed method were evaluated.'' 
\\\hline
 {\bf T5-base:} ``Fuzzy based self organizing aggregation method was proposed for swarm robots. Swarm robots evaluated their limited sensor input via rules of fuzzy logic. Swarm robots displayed aggregation behavior during the systematic experiments applied. Swarm robots displayed aggregation behavior during the systematic experiments applied.''\\\hline 	    
 {\bf BART-base:} ``Fuzzy based self organizing aggregation method was proposed for swarm robots. Self organizing aggregation behavior was evaluated on different number of swarm robots with different detection areas in arenas of different sizes. The performance of the fuzzy logic based method was evaluated by systematic experiments. The scalability and flexibility of the self organizing method were evaluated by way of systematic experiments.''\\\hline
 {\bf LLaMA-3-8B:} ``A novel Fuzzy Logic Based Self Organizing Aggregation Method is proposed. This method enables swarm robots to aggregate without any need for communication. It can be used as a building block for designing more complex systems. It has been tested under various conditions including noisy environments. The results show its robustness and efficiency.''\\\hline
 \end{tabular} 	
 \caption{\small Highlights generated by the different fine-tuned models are shown.  The input is an abstract from the \textbf{MixSub} test split, sourced from  \texttt{\small \url{https://www.sciencedirect.com/science/article/pii/S0303264720300836}}.}	
 \label{fig:sample_Abs_Highlights_MixSub} 
 \end{figure*}

\subsection{Analysis of Energy Consumption}
\begin{figure*}[!htbp] 
\centering
\includegraphics[width=0.98\linewidth]{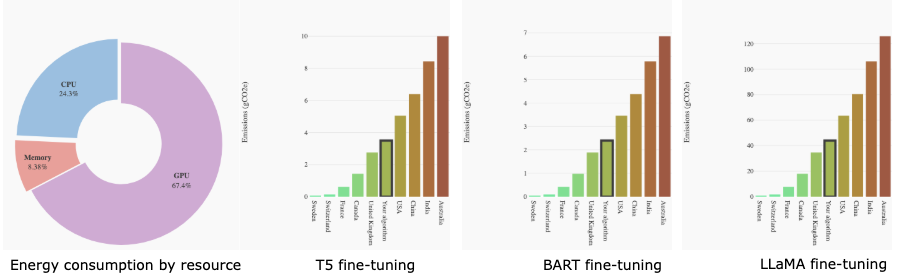}
\caption{\small Energy consumption by resource (GPU, CPU and memory), and carbon footprints for model fine-tuning for one epoch. The resource-wise shares of energy consumption (leftmost figure) are identical for all our models (GPU: 67.4\%, CPU: 24.3\%, memory: 8.3\%). The three figures on the right show carbon footprints (in gCO$_{2}$e) on the Y-axis for different locations of the data centers on the X-axis, and highlight the location we used. The carbon footprints are 3.5 gCO$_{2}$e for T5-base, 2.4 gCO$_{2}$e for BART-base,  and 43.98 gCO$_{2}$e for LLaMA-3-8B.}
\label{fig:model_diagram}
\end{figure*}

We use Equation \ref{eq:Our_carbon_footprint}, as implemented in the Green Algorithms (GA) calculator\footnote{\url{https://calculator.green-algorithms.org/}}, to evaluate the carbon footprint of the fine-tuning operations we performed for the models. All our models were trained on the Nvidia Tesla A100 40GB SXM4 GPU provided by \texttt{Colab Pro+}, which is equipped with GPU support. Google Colab does not reveal the exact CPU information except that it is a Intel Xeon processor with 12 cores, so we use the option `Xeon E5-2680 v3' (which comprises 12 cores) in GA. The memory is 83.5 GB RAM (as observed with \texttt{!cat /proc/meminfo}). In GA, the total amount of available memory, rather than the memory actually consumed, is considered. The data center used is located in Iowa, USA. According to \cite{PUE}, Google utilizes machine learning techniques to keep its global average PUE to 1.10, compared to the industry average of 1.58. We assumed 100\% CPU and GPU utilization, to simplify the computation, and so this is a slight overestimation. We observed that the time taken to execute 1 epoch is  1.75 mins for T5-base, 1.2 mins for BART-base, and 22 mins for LLaMA-3-8B. With these values, the GA calculator gives epoch-wise carbon footprint as \textbf{3.5} gCO$_2$e (energy consumed 11.91 Wh) for \textbf{T5-base}, \textbf{2.4} gCO$_{2}$e (8.16 Wh) for \textbf{BART-base}, and \textbf{43.98} gCO$_{2}$e (149.68 Wh) for \textbf{LLaMA-3-8B}, as depicted in Figure \ref{fig:model_diagram}. Fine-tuning LLaMA for one epoch has a carbon footprint equivalent to 0.09\% of CO$_{2}$ emission from a Paris-London flight, or 0.02\% of a Kolkata-Dehradun flight (assuming around 139 gCO$_2$e per passenger km). LLaMA-3-8B's significantly larger parameter count explains its higher carbon emissions, both per epoch and over the entire fine-tuning duration.

\begin{figure}[!htbp]
    \centering
    \subfloat{\includegraphics[width=0.48\textwidth, keepaspectratio]{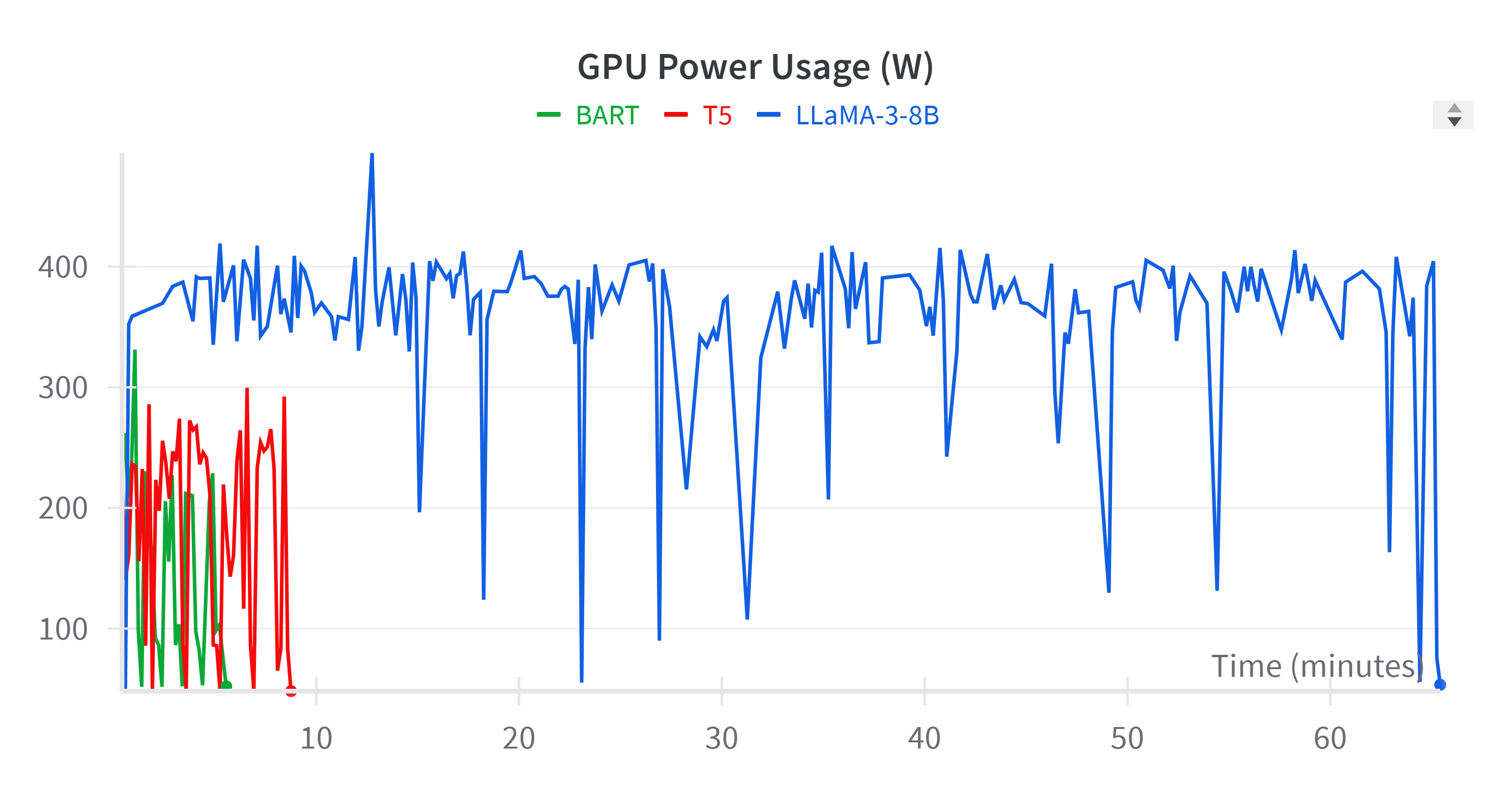}}\hspace{0.1cm}
    \subfloat{\includegraphics[width=0.48\textwidth, keepaspectratio]{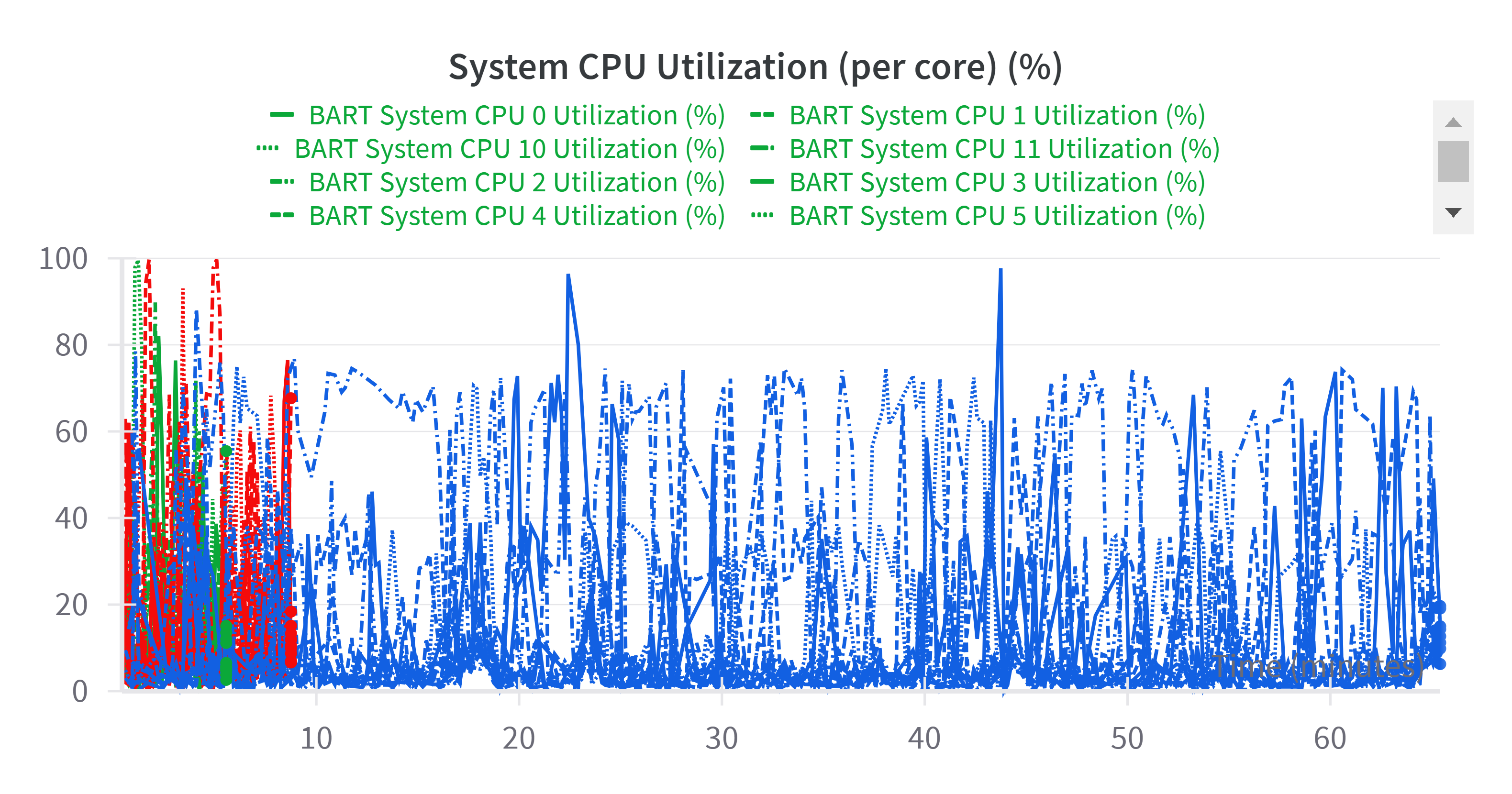}}\hspace{0.1cm}
    \subfloat{\includegraphics[width=0.48\textwidth, keepaspectratio]{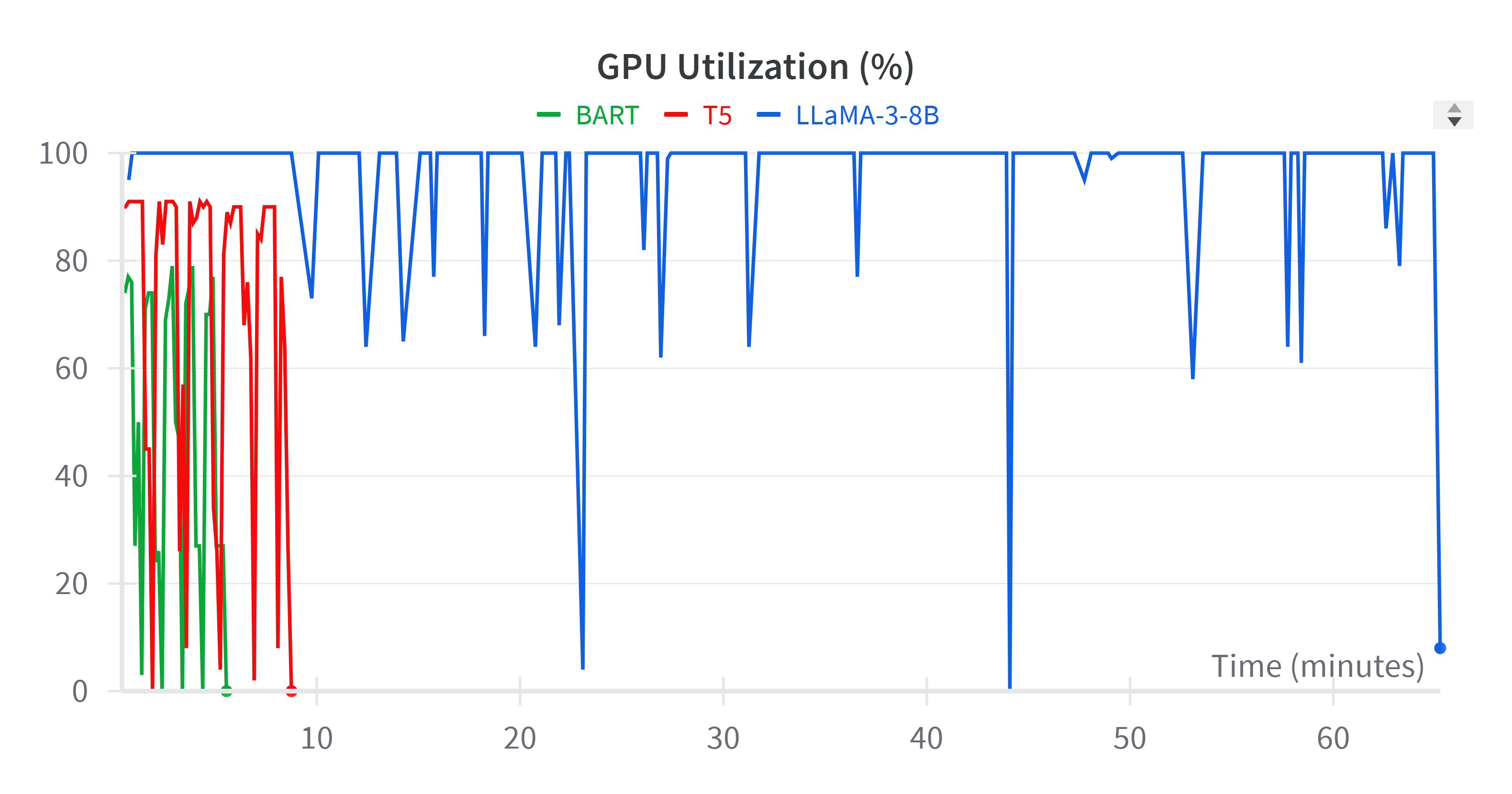}}\hspace{0.1cm}
    \subfloat{\includegraphics[width=0.48\textwidth, keepaspectratio]{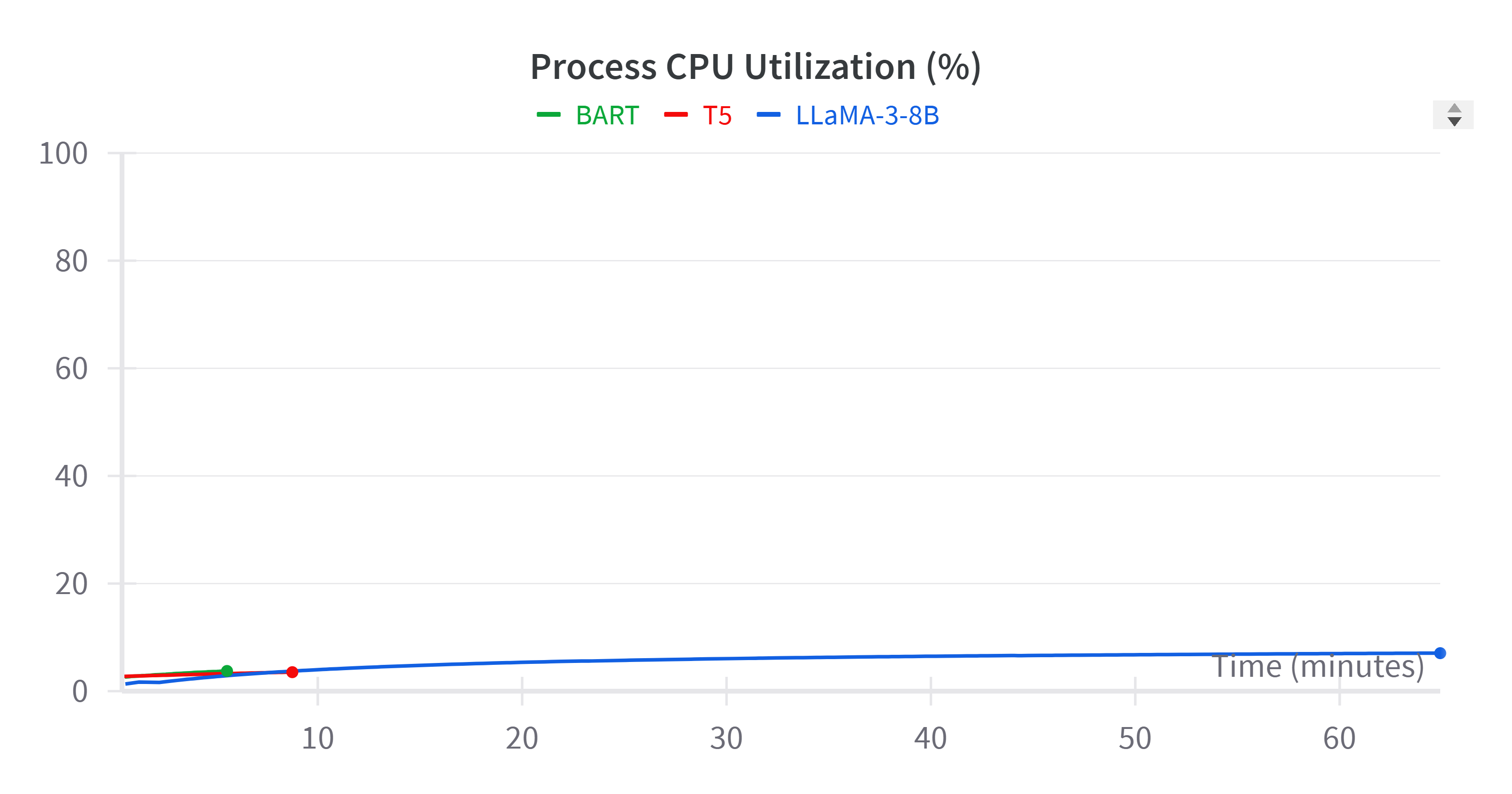}}\hspace{0.1cm}
     \subfloat{\includegraphics[width=0.48\textwidth, keepaspectratio]{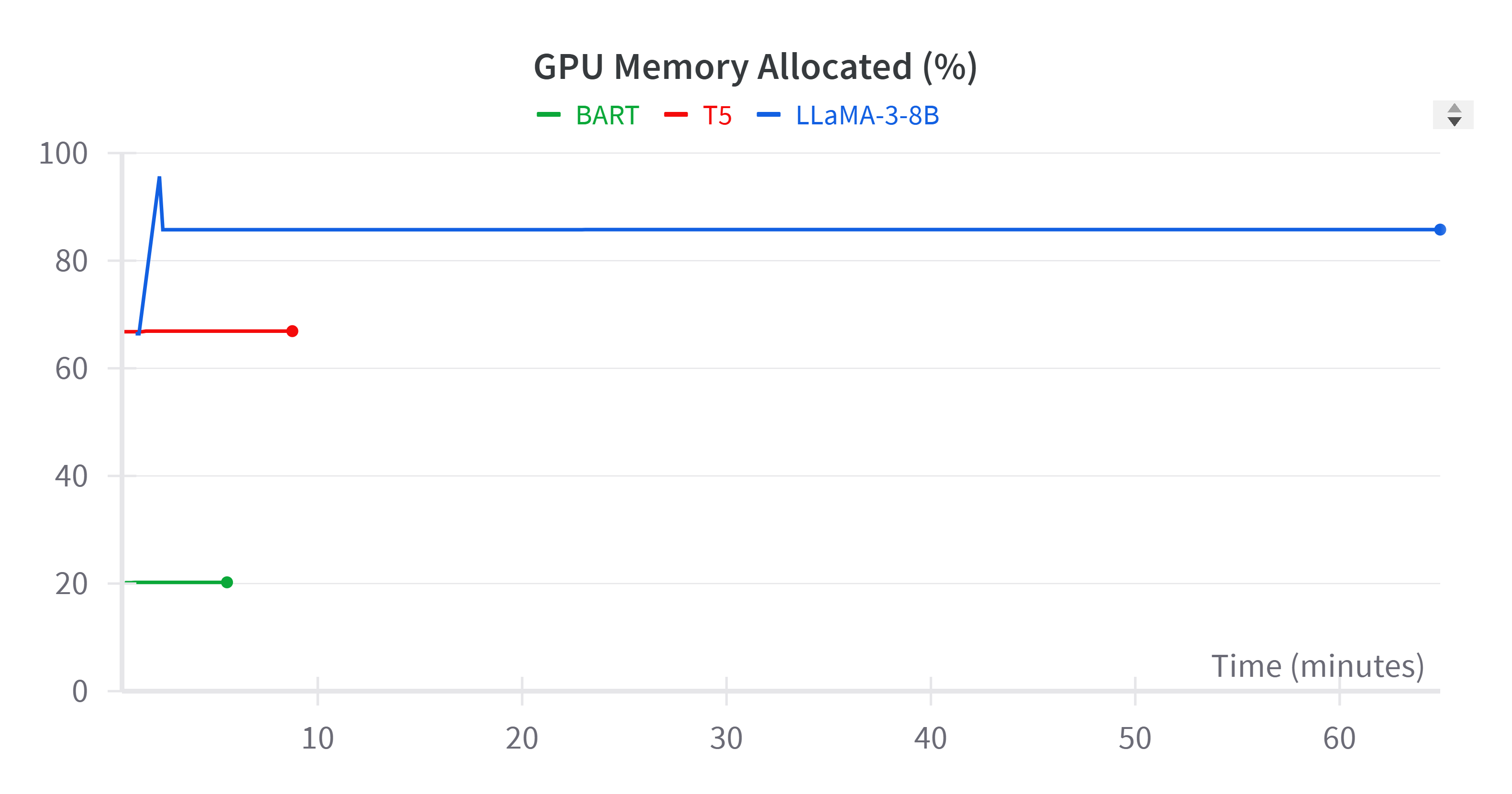}}\hspace{0.1cm}
    \subfloat{\includegraphics[width=0.48\textwidth, keepaspectratio]{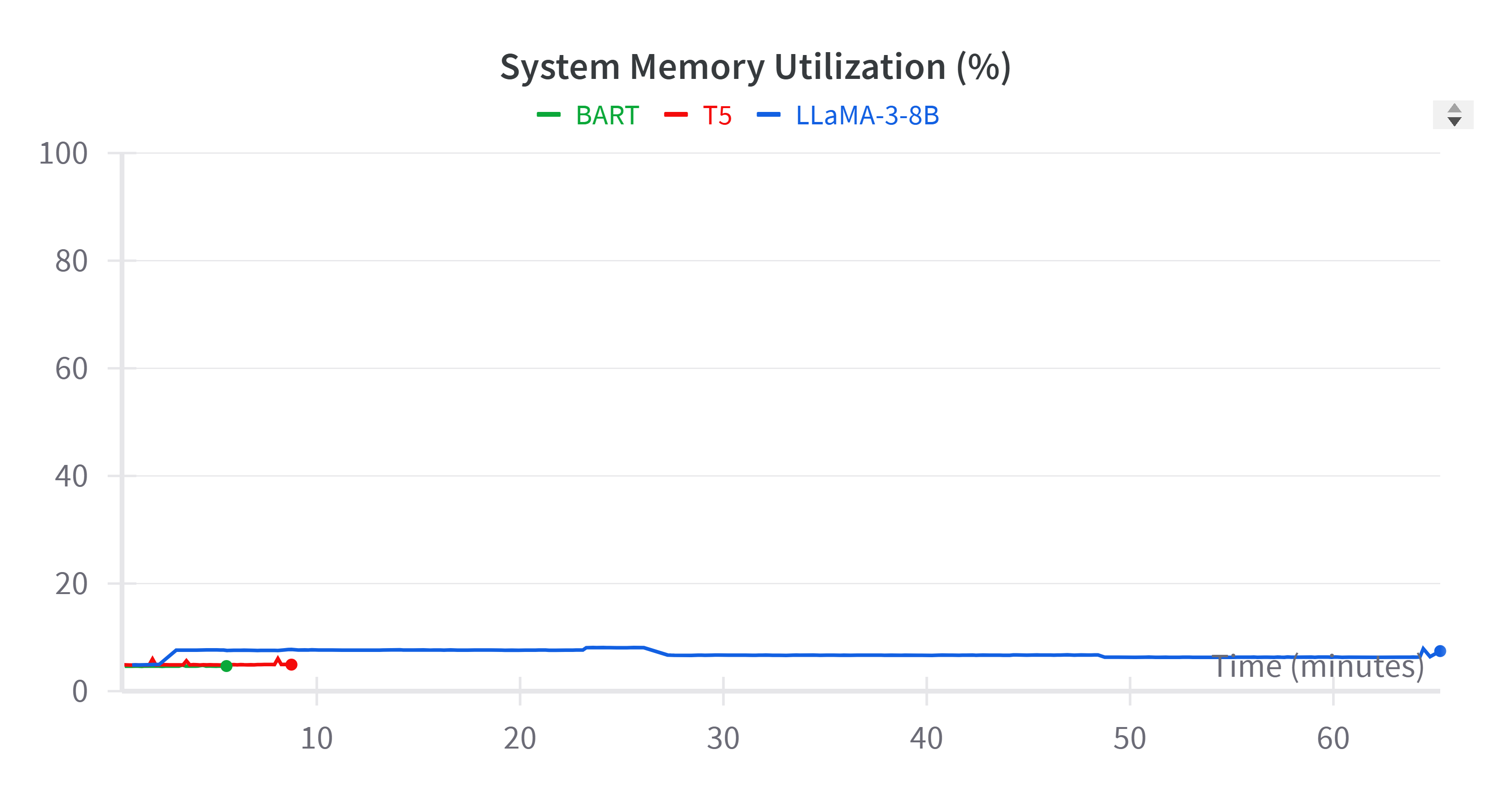}}\hspace{0.1cm}  
    \caption{\small Comparison of computational resources utilized during fine-tuning of the PLMs (T5-base and BART-base) and LLaMA-3-8B LLM for the text summarization task.}
    \label{fig:resource_ts}
\end{figure}

We have used the WandB tool\footnote{\url{https://wandb.ai/site}} to track the consumption of memory and compute resources during the fine-tuning process. As visualized in Figure \ref{fig:resource_ts}, GPU and CPU usage values are time-dependent. Clearly, LLaMA-3-8B uses close to 100\% GPU most of the time while it is much lower for the other two models. The GPU memory allocation  and system memory allocation are also significantly higher for LLaMA-3-8B. This clearly shows the huge resource, energy and environmental costs associated with even fine-tuning models for NLP tasks.


\section{Conclusion}
We have evaluated the performance and environmental impact of three pre-trained language models: LLaMA 3-8B, T5, and BART. Although LLaMA 3-8B produced lower ROUGE and METEOR scores than the other two models, it achieved comparable performance in terms of semantic similarity. However, its environmental cost was significantly higher.  This shows that smaller fine-tuned models remain effective for summarization tasks. Future research should prioritize reducing the carbon footprint of language models.

\subsubsection{Acknowledgements.} This work is supported by the project  CRG/2021/000803, funded by the Department of Science and Technology, Government of India.

\bibliographystyle{plain}
\bibliography{anthology}
\end{document}